\documentclass[letterpaper]{article} 
\usepackage{aaai24}  
\usepackage{times}  
\usepackage{helvet}  
\usepackage{courier}  
\usepackage[hyphens]{url}  
\usepackage{graphicx} 
\usepackage{natbib}  
\usepackage{caption} 
\usepackage{graphicx} 
\usepackage{epsfig}
\usepackage{float} 
\usepackage{subfigure} 
\usepackage{color}
\usepackage{booktabs}
\usepackage{amsmath}
\usepackage{amssymb}
\usepackage{multirow}
\usepackage{algorithm}
\usepackage{algorithmic}
\newcommand{\ie}{\textit{i}.\textit{e}.}

\usepackage{newfloat}
\usepackage{listings}
\DeclareCaptionStyle{ruled}{labelfont=normalfont,labelsep=colon,strut=off} 
\floatstyle{ruled}
\newfloat{listing}{tb}{lst}{}
\floatname{listing}{Listing}
\title{SimDistill: Simulated Multi-modal Distillation for BEV 3D Object Detection}
\author{
    Haimei Zhao\textsuperscript{\rm 1},
    Qiming Zhang\textsuperscript{\rm 1},
    Shanshan Zhao\textsuperscript{\rm 1}
    Zhe Chen\textsuperscript{\rm 2}
    Jing Zhang\textsuperscript{\rm 1},
    Dacheng Tao\textsuperscript{\rm 1}
}
\affiliations{
    \textsuperscript{\rm 1}School of Computer Science, The University of Sydney, Australia, \\
    \textsuperscript{\rm 2}School of Computing, Engineering and Mathematical Sciences, La Trobe University, Australia\\

    hzha7798@uni.sydney.edu.au, qzha2506@uni.sydney.edu.au,
    sshan.zhao00@gmail.com,
    zhe.chen@latrobe.edu.au,
    jing.zhang1@sydney.edu.au,
    dacheng.tao@gmail.com
}

\usepackage{bibentry}

\begin{document}

\maketitle

\begin{abstract}
Multi-view camera-based 3D object detection has become popular due to its low cost, but accurately inferring 3D geometry solely from camera data remains challenging and may lead to inferior performance. Although distilling precise 3D geometry knowledge from LiDAR data could help tackle this challenge, the benefits of LiDAR information could be greatly hindered by the significant modality gap between different sensory modalities. To address this issue, we propose a \textbf{Si}mulated \textbf{m}ulti-modal \textbf{Distill}ation (\textbf{SimDistill}) method by carefully crafting the model architecture and distillation strategy. Specifically, we devise multi-modal architectures for both teacher and student models, including a LiDAR-camera fusion-based teacher and a simulated fusion-based student. Owing to the ``identical'' architecture design, the student can mimic the teacher to generate multi-modal features with merely multi-view images as input, where a geometry compensation module is introduced to bridge the modality gap. Furthermore, we propose a comprehensive multi-modal distillation scheme that supports intra-modal, cross-modal, and multi-modal fusion distillation simultaneously in the Bird's-eye-view space. Incorporating them together, our SimDistill can learn better feature representations for 3D object detection while maintaining a cost-effective camera-only deployment. Extensive experiments validate the effectiveness and superiority of SimDistill over state-of-the-art methods, achieving an improvement of 4.8\% mAP and 4.1\% NDS over the baseline detector. The source code will be released at https://github.com/ViTAE-Transformer/SimDistill.
\end{abstract}

\section{Introduction}

\begin{figure}[t]
\centering
\includegraphics[width=0.9\linewidth]{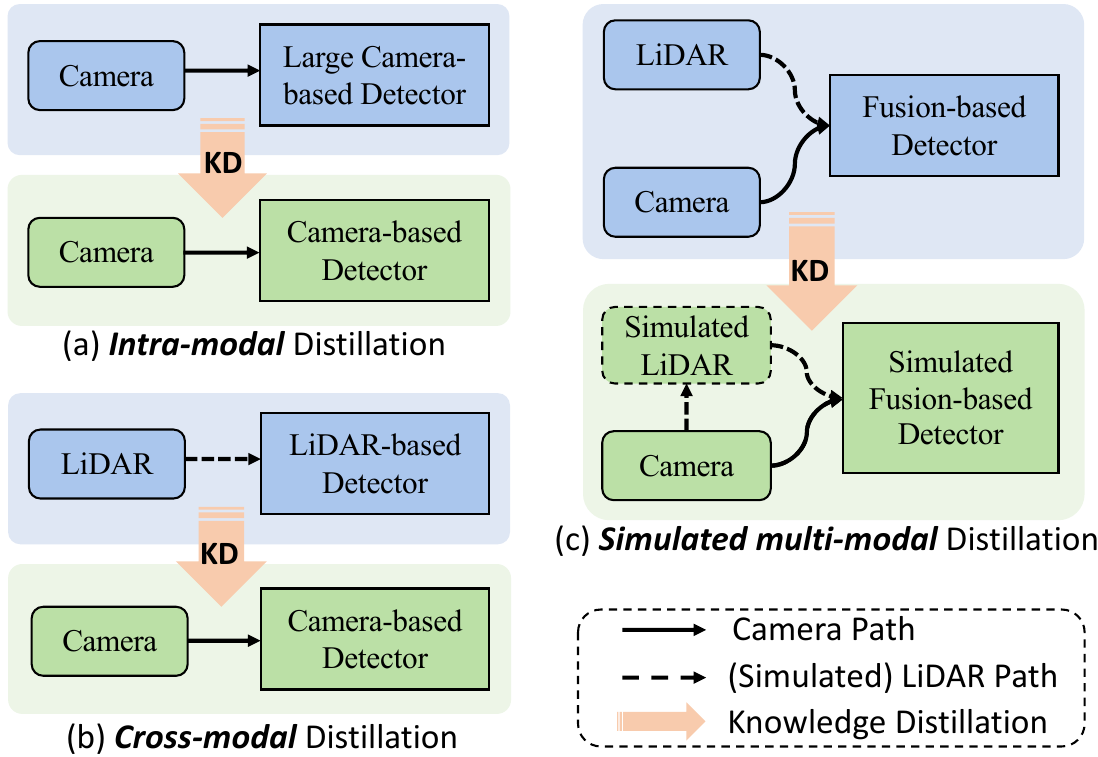}
\caption{Comparison of our SimDistill with previous distillation frameworks. (a) Intra-modal distillation between camera-only teacher and student models cannot learn accurate 3D information due to the limited capacity of the teacher model for inferring 3D geometry. (b) Cross-modal distillation between the LiDAR teacher and Camera student enables learning useful 3D information from the teacher but suffers from the large cross-modal gap. (c) Our simulated multi-modal distillation enables effective knowledge distillation within/between modalities and fully takes advantage of complementary information from different modalities.}
\label{fig:startpic}
\end{figure}

3D object detection is a pivotal technique with extensive applications in fields such as autonomous driving, robotics, and virtual/augmented reality \cite{zhang2020empowering}. In recent years, camera-based 3D object detection methods, which infer objects' 3D locations from multi-view images \cite{huang2021bevdet,li2022bevdepth}, have attracted great attention from both academia and industry because of the high perceptual ability of dense color and texture information with low deployment cost. However, due to the lack of accurate 3D geometry reasoning ability, their detection performance falls largely behind LiDAR-based methods, which poses a challenge to the practical deployment of camera-based methods.

To address this issue, researchers attempt to impose LiDAR data to provide accurate 3D geometry information. Some multi-view camera-based methods \cite{li2022bevdepth,li2023bevstereo} generate ground truth depth from LiDAR point cloud and use it as the supervisory signal for depth estimation to help transform image features to the Bird's-eye-view (BEV) space \cite{zhao2022jperceiver} accurately. Except for directly using LiDAR as supervision during training, some recent work employs LiDAR information by applying the knowledge distillation (KD) technique~\cite{gou2021knowledge} to improve the detection performance of camera-based methods.
KD-based 3D object detection methods usually leverage the informative features or predictions of a well-trained teacher model to facilitate the learning of the student model. One straightforward approach is intra-modal distillation \cite{li2022bev-lgkd,zhang2022structured} between a large teacher model and a small student model, as shown in Figure~\ref{fig:startpic} \textbf{(a)}, which conducts distillation within the image modality.
However, the ceiling performance of the model can be limited since the teacher model infers 3D geometry solely from image data as well. Another approach is cross-modal distillation, as shown in Figure~\ref{fig:startpic} \textbf{(b)}, which utilizes LiDAR data as the input of teacher models and transfers 3D knowledge to camera-based students~\cite{chong2021monodistill,chen2023bevdistill,li2022unifying}. The student is usually forced to learn and mimic the output of a LiDAR-based teacher in different representation spaces, including monocular view features \cite{chong2021monodistill}, BEV features \cite{chen2023bevdistill}, and voxel features~\cite{li2022unifying}. Nevertheless, performing knowledge distillation directly between different modalities might face significant cross-modal gaps and struggle in aligning features learned by distinct architectures of teacher and student models, resulting in limited performance improvements.

In this paper, we address this challenge from the perspective of architecture design and multi-modal knowledge distillation scheme, presenting a \textbf{Si}mulated \textbf{m}ulti-modal \textbf{Distill}ation (\textbf{SimDistill}) method for 3D object detection. 
It encourages the student to \emph{\textbf{simulate}} multi-modal representation with solely image modality as input thereby advancing the representation learning for 3D object detection. 
For the architecture, we design a LiDAR-camera fusion-based teacher and a simulated multi-modal student. The student model not only involves a camera path but also introduces an additional simulated LiDAR path parallel to the camera counterpart, as shown in Figure \ref{fig:startpic} \textbf{(c)}. Different from other distillation methods in Figure \ref{fig:startpic} \textbf{(a)} and \textbf{(b)}, our student model possesses two knowledge-transferring paths to learn complementary information from the corresponding two branches of the teacher model. Despite the simulation nature, our student shares a nearly ``identical'' pipeline as the teacher to produce the camera feature, LiDAR feature, fusion feature, and detection predictions. The resulting aligned learning workflow greatly mitigates the cross-modal gap and benefits multi-modal knowledge distillation.

Built upon this architecture, we propose a new simulated multi-modal distillation scheme that supports intra-modal (IMD), cross-modal (CMD), and multi-modal fusion distillation (MMD) simultaneously. We adopt the widely used MSE loss on the corresponding feature representations distillation in the unified BEV space and an additional quality-aware prediction distillation \cite{hong2022cross}. It is noteworthy that directly transferring knowledge from the LiDAR feature to the simulated LiDAR feature is challenging due to the cross-modal gap. To approach this challenge, we devise a geometry compensation module in CMD to help it attend more to the valuable surrounding context from the learned locations to conduct geometry remediation and distill more informative features from object regions. Equipping the proposed model with the distillation scheme, our SimDistill could effectively learn better feature representations for 3D object detection while enjoying cost-effective camera-only deployment.

The main contribution of this paper is threefold. \textbf{Firstly}, we propose a unique multi-modal distillation framework for BEV 3D object detection, including a LiDAR-camera fusion-based teacher and a carefully crafted simulated multi-modal student. By ensuring that the teacher and student models share nearly the same workflows, we effectively reduce the modality gap in knowledge distillation. \textbf{Secondly}, we present a novel simulated multi-modal distillation scheme that supports intra-modal, cross-modal, and multi-modal fusion distillation simultaneously, which is a universal strategy and can be easily adapted to different models. \textbf{Thirdly}, comprehensive experiments and ablation studies on the nuScenes benchmark validate the effectiveness of SimDistill and its superiority over existing state-of-the-art methods, improving the mAP and NDS of the baseline detector by 4.8\% and 4.1\%, respectively. 

\section{Related Work}

\paragraph{Camera-based 3D Object Detection}
Monocular 3D object detection methods have been widely studied and made great progress \cite{simonelli2019disentangling,reading2021categorical,wang2021fcos3d,lu2021geometry,ma2021delving,huang2022monodtr} on the KITTI \cite{geiger2012we} benchmark. However, with the release of large-scale datasets with multi-view cameras such as nuScenes \cite{caesar2020nuscenes} and Waymo \cite{sun2020scalability}, there is growing attention for accurate 3D object detection in these more challenging scenes.
Recent works adopt the Bird's-eye view (BEV) representation as an ideal feature space for multi-view perception due to its excellent ability to address scale-ambiguity and occlusion issues \cite{huang2021bevdet,huang2022bevdet4d,li2022bevformer}. Various methods have been proposed to transform perspective image features to the BEV space, such as the lifting operation from LSS \cite{philion2020lift} used by BEVDet \cite{huang2021bevdet} and the cross-attention mechanism-based grid queries used by BEVFormer \cite{li2022bevformer}. The camera-based BEVDet approach has been further improved by imposing depth supervision \cite{li2022bevdepth,li2023bevstereo,wang2022sts,chu2023oa} and temporal aggregation \cite{huang2022bevdet4d,park2022time}, resulting in better performance. However, there is still a significant performance gap compared to LiDAR-based and fusion-based counterparts.

 \paragraph{Fusion-based 3D Object Detection}
LiDAR differs from cameras in its ability to capture precise geometric and structural information. However, the data it produces is sparse and irregular, with a large volume. Some methods use PointNet \cite{qi2017pointnet} directly on the raw point cloud \cite{qi2017pointnet++,shi2019pointrcnn,chen2022sasa} to learn 3D features, while others voxelize the point cloud into pillars \cite{lang2019pointpillars,wang2020pillar,yin2021center} or voxels \cite{zhou2018voxelnet,yan2018second} before extracting features using SparseConvNet \cite{graham20183d}. State-of-the-art techniques \cite{yin2021center,bai2022transfusion} typically transform 3D features into the BEV representation to simplify operations in 3D space, and then feed the resultant features to subsequent detection heads.

Due to their distinct strengths in perceiving, both cameras and LiDAR are integrated into sensor fusion methods to enhance the performance of perception systems. Existing fusion-based approaches can be categorized as input-level methods \cite{vora2020pointpainting,wang2021pointaugmenting,xu2021fusionpainting} and feature-level methods \cite{bai2022transfusion,liang2022bevfusion,liu2022bevfusion,yan2023cross}, depending on the stage at which information from different sensors is combined. Recently, it has been shown that BEV space is an ideal space for multi-modal fusion, resulting in outstanding performance \cite{liang2022bevfusion,liu2022bevfusion}. 
These methods follow a simple yet effective pipeline that involves extracting features from both modalities, transforming features into the BEV space, fusing multi-modal features using fusion modules, and conducting subsequent detection, largely improving the performance.
 
\begin{figure*}[t]
\begin{center}
  \includegraphics[width=0.92\linewidth]{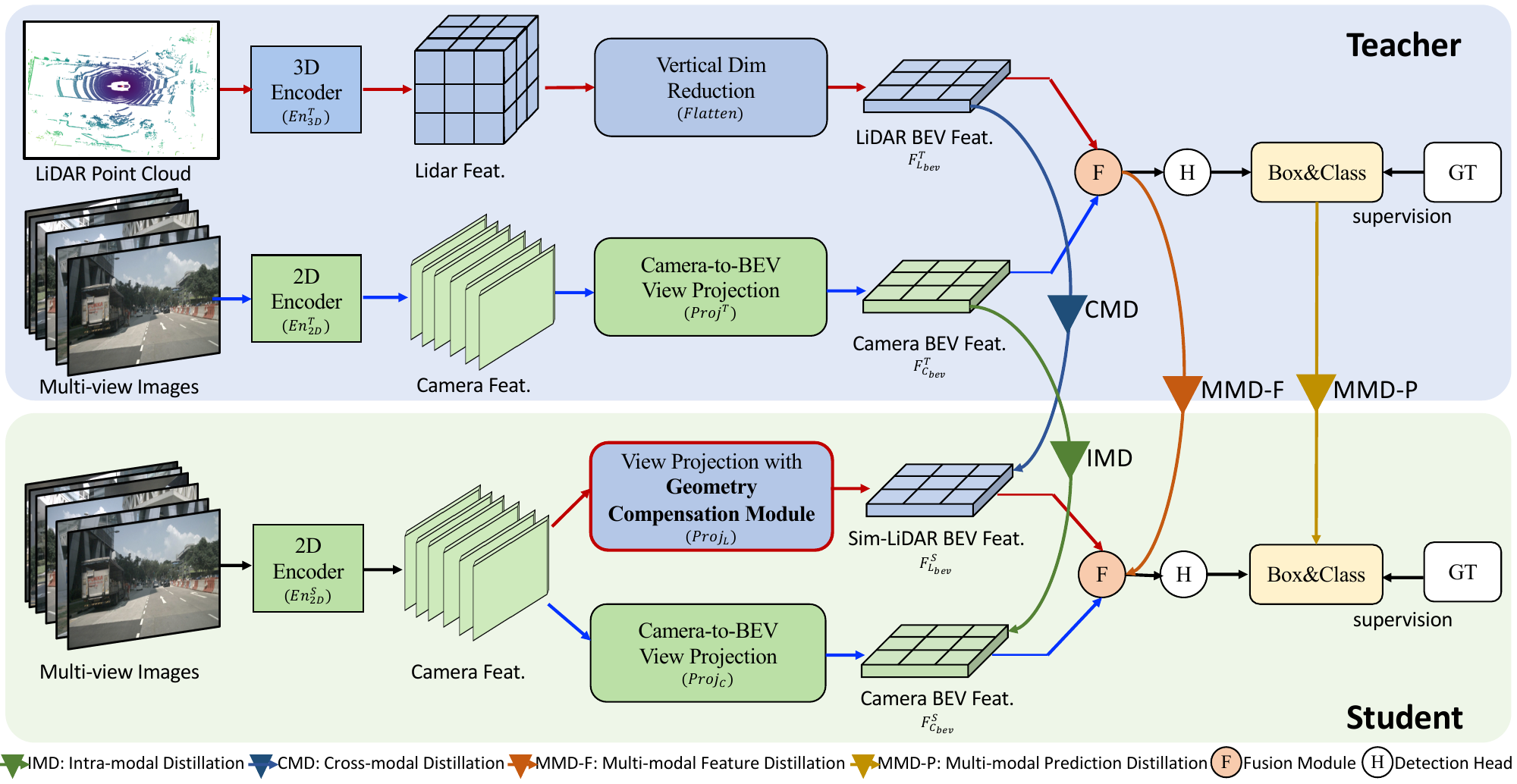}
\end{center}
  \caption{Overall pipeline of SimDistill. It consists of a fusion-based teacher model (top) and a simulated multi-modal student model (bottom). SimDistill supports (1) Intra-Modal Distillation (IMD) between the camera features of the teacher and student; (2) Cross-Modal Distillation (CMD) between the teacher's LiDAR feature and the student's Simulated-LiDAR feature. 
  (3) Multi-Modal fusion Distillation (MMD) between the fusion features (MMD-F) and predictions (MMD-P) of the teacher and student. The workflows of the (simulated) LiDAR and camera branches are denoted by red and blue arrows, respectively.}
\label{fig:mainfig}
\end{figure*}

\paragraph{Knowledge Distillation in 3D Object Detection}
Knowledge distillation presents a promising avenue for empowering compact models (\ie, students) with effective representations via knowledge transfer from larger models (\ie, teachers). In the context of 3D object detection, prior research \cite{zhang2022structured,yang2022towards,cho2023itkd} has successfully extended knowledge distillation techniques, requiring the student network to emulate features or predictions learned by a teacher model within the same modality. Recent advancements in the area of KD-based 3D object detection have ventured into employing teachers from different modalities \cite{chong2021monodistill,li2022bev-lgkd,hong2022cross,chen2023bevdistill}, \ie, leveraging a LiDAR-based teacher. UVTR \cite{li2022unifying} aligns features from both LiDAR and camera in voxel space, facilitating knowledge distillation. BEVDistill \cite{chen2023bevdistill} transforms features into the BEV space for the feature and instance-wise prediction distillation. In a similar vein, TiG-BEV \cite{huang2022tig} introduces inner-depth supervision and inner-feature distillation to enhance geometry learning in the BEV space. These cross-modal distillation techniques underscore the potential of transferring knowledge from robust LiDAR teachers to camera-based students. Nevertheless, 
these approaches overlook the prospect of distilling multi-modal knowledge for 3D object detection.
Our approach diverges by exploring a multi-modal teacher and designing a nearly identical yet simulated multi-modal architecture alongside tailored distillation schemes to effectively perform multi-modal distillation. While concurrent work Unidistill \cite{zhou2023unidistill} also embraces a multi-modal teacher, it is designed as a universal knowledge distillation framework to support both single-to-single and fusion-to-single cross-modal distillation. It pays no attention to the architecture discrepancy issue between teacher and student and fails to perform comprehensive multi-modal distillation and overcome the cross-modal gap.

\section{Methodology}
In this section, we present the details of how the proposed SimDistill realizes comprehensive multi-modal knowledge distillation for 3D object detection. We first introduce the model architecture, which consists of a multi-modal fusion-based teacher and a simulated multi-modal student. Next, we describe the simulated multi-modal distillation scheme that supports knowledge distillation within and between modalities. Last, we present the training objectives for our method.

\subsection{Multi-modal Architecture}\label{sec:structure}
SimDistill is proposed as a flexible multi-modal distillation method, offering the flexibility to select both the teacher model and the student model from diverse methods. In the subsequent sections, we present a concrete implementation of SimDistill, employing BEVFusion \cite{liu2022bevfusion} as the teacher model and design the student model based on the camera branch of BEVFusion (BEVFusion-C). The architectural layout of SimDistill is depicted in Figure \ref{fig:mainfig}. The upper block depicts the configuration of the teacher model, while the lower block represents the student model. In both instances, the LiDAR branch and the camera branch workflows are denoted by red and blue arrows, respectively.
\paragraph{Multi-modal Teacher}\label{sec:teacher}
To encode multi-modal knowledge effectively, we adopt the state-of-the-art fusion-based method, \ie, BEVFusion \cite{liu2022bevfusion} as the teacher model. Its architecture comprises two branches, as depicted in the top part of Figure~\ref{fig:mainfig}. The LiDAR branch follows the standard pipeline of a LiDAR-based detector \cite{yan2018second,yin2021center}. It uses SparseConvNet \cite{graham20183d} $En^T_{3D}$ to extract the 3D features, and obtains the BEV features $F^{T}_{L_{bev}}$ through vertical dimension reduction ($\rm Flatten$). On the other hand, the camera branch follows the paradigm of BEVDet \cite{huang2021bevdet}, using a 2D feature extractor $En^T_{2D}$ and an efficient projection $Proj^T$ to transform features from the camera view to the BEV space $F^{T}_{C_{bev}}$. Both modalities' features are then embedded in a unified BEV space using a fully-convolutional fusion module $fuse^T$, which produces the fused BEV features $F^{T}_{U_{bev}}$. Finally, a detection head $head^T$ predicts the objects' bounding boxes and classes $P^{T}$. This process is formulated as:
\begin{equation}
    \begin{split}
        F^{T}_{L_{bev}} &= ~{\rm Flatten}(En^T_{3D}(L)),\\
        F^{T}_{C_{bev}} &= ~Proj^T(En^T_{2D}(I)),\\
        F^{T}_{U_{bev}} &= ~fuse^T(F^{T}_{L_{bev}}, F^{T}_{C_{bev}}),\\
        P^{T} &= ~head^T(F^{T}_{U_{bev}}),
    \end{split}
\end{equation} where $L$ and $I$ denotes LiDAR and image input. $T$ and $S$ in all formulations represent the teacher and student models. The projection $Proj$ will be explained in the following part.

\paragraph{Simulated multi-modal Student}\label{sec:student}
For the student model, we adopt BEVFusion-C \cite{liu2022bevfusion} as the basis model. To mimic the multi-modal fusion pipeline of the teacher model, we make a modification to the network, as shown in the bottom part of Figure~\ref{fig:mainfig}. Specifically, after feature extraction from the 2D encoder $En^S_{2D}$, we devise an additional simulated LiDAR branch (workflow denoted with red arrows) in parallel to the camera branch (blue arrows in the bottom) to simulate LiDAR features from images, which are supervised by the real LiDAR features from the teacher.

In the camera branch, we adopt the same efficient view projection $Proj_C$ with the one used in the teacher model ($Proj^T$) to transform camera-view features to the corresponding BEV features $F^{S}_{C_{bev}}$ \cite{philion2020lift,liu2022bevfusion}. During the feature transformation, the extracted 2D feature $F^{S}_{C_{uv}}$ is first feed to a light Depth Net $\phi$ and a Context Net $\psi$ to predict the depth distribution and semantic context on each pixel. Then, each 2D feature pixel can be scattered into $D$ discrete points along the camera ray by rescaling the context feature with their corresponding depth probabilities. The resulting 3D feature point cloud is then processed by the efficient BEV pooling operation $\rho$, to aggregate features in BEV grids and obtain the BEV features:
\begin{equation}
F^{S}_{C_{bev}}= Proj_C(F^{S}_{C_{uv}})  = \rho(\psi(F^{S}_{C_{uv}}) \times \phi(F^{S}_{C_{uv}})).
\label{eq:bevfeature}
\end{equation}

In the simulated LiDAR branch, to acquire the simulated LiDAR feature $F^{S}_{L_{bev}}$, the view projection $Proj_L$ is combined with a specifically designed geometry compensation module in both camera-view and BEV spaces, which will be explained later in Eq.~\eqref{eq:gcm} of Sec. 3.2.2. It offers the ability to mitigate the geometry misalignment caused by inaccurate depth prediction and modality gap during distillation. After obtaining BEV features from two branches $F^{S}_{C_{bev}}$ and $F^{S}_{L_{bev}}$, we use the fusion module $fuse^S$ to acquire the multi-modal fusion features $F^S_{U_{bev}}$. And the detection head $head^S$ is exploited to yield the final detection results $P^{S}$. Both the fusion module and detection head have the same architecture as the teacher. This process is formulated as:
\begin{equation}
    \begin{split}
    F^{S}_{C_{uv}} = ~ &En^S_{2D}(I),\\
    F^{S}_{L_{bev}} = Proj_L(F^{S}_{C_{uv}}),&\quad
    F^{S}_{C_{bev}} = Proj_C(F^{S}_{C_{uv}}),\\
    F^S_{U_{bev}} = ~ &fuse^S(F^{S}_{L_{bev}}, F^{S}_{C_{bev}}),\\
    P^{S} = ~ &head^S(F^S_{U_{bev}}).
    \end{split}
\end{equation}
Owing to the simulated multi-modal fusion architecture, the student model can learn features from multiple modalities without equipping a real LiDAR. In the next part, we will explain how this architecture facilitates effective knowledge distillation within and between modalities, including intra-modal, cross-modal, and multi-modal fusion distillation.

\subsection{Multi-modal Distillation }
\label{sec:simMMD}
To better utilize the knowledge of different modalities encoded by different branches of the teacher model, we propose a novel simulated multi-modal distillation scheme including Intra-modal Distillation (IMD), Cross-modal Distillation (CMD), and Multi-modal fusion Distillation (MMD).
\subsubsection{Intra-modal Distillation}\label{sec:IMD}
Since both the teacher and student models take images as input, a straightforward strategy is to align the image features from the camera branch of both models, which we name intra-modal distillation.
Specifically, we leverage the BEV feature of the teacher $F^{T}_{C_{bev}}$ as the supervisory signal for the learning of the student counterpart $F^{S}_{C_{bev}}$ via an MSE loss, \ie,
\begin{equation}
     \mathcal L_{IMD} = {\rm MSE}(F^T_{C_{bev}}, F^S_{C_{bev}}).
\end{equation}
Due to the same modality in IMD, the student model can be trained directly through the above distillation objective to gain useful visual domain knowledge to facilitate 3D object detection performance. However, relying on images alone may not provide enough geometry-related information to help detect target objects. To address this limitation, we implement cross-modal distillation on the proposed simulated LiDAR branch in the student model, enabling it to gain knowledge from the LiDAR modality.
\begin{figure}[t]
\centering
\includegraphics[width=1\linewidth]{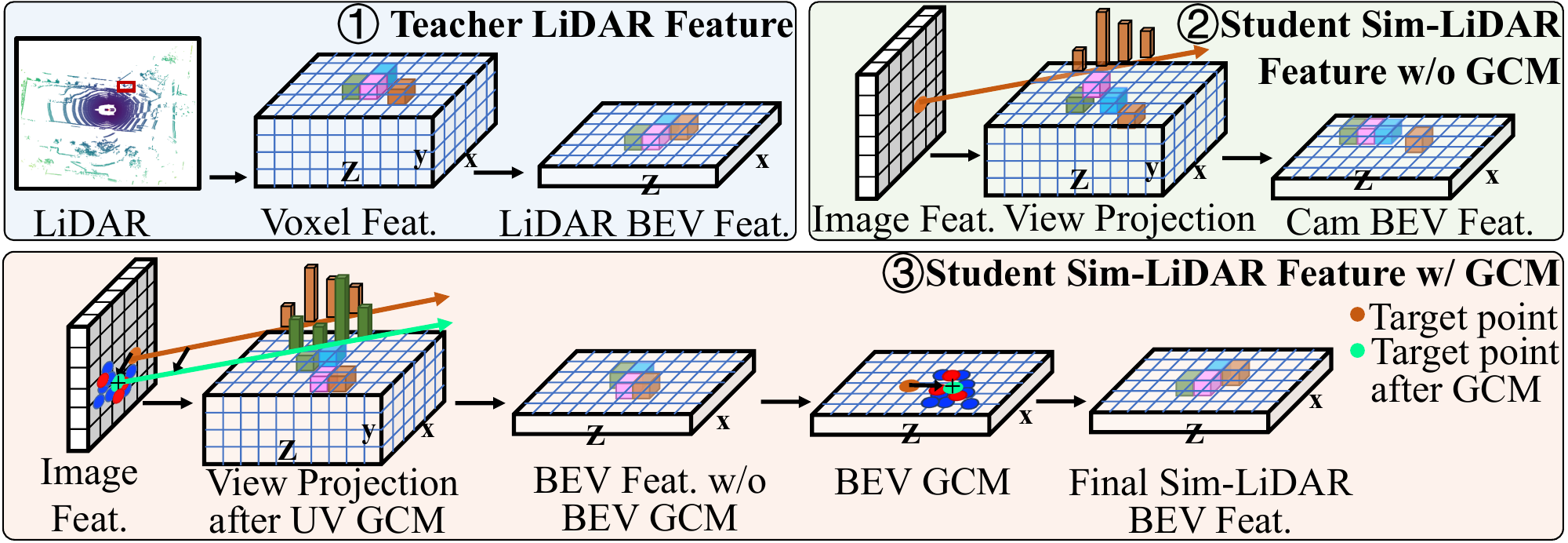}
\caption{Illustration of Geometry Compensation Module (GCM). The colorful voxels denote learned features of the target object. Best viewed with zoom-in.}
\label{fig:GCMillustration}
\end{figure}
\subsubsection{Cross-modal Distillation}\label{sec:CMD}
CMD aims to align the LiDAR BEV features of the teacher and the simulated LiDAR BEV features of the student. However, due to geometry misalignment and modal difference, directly applying the distillation loss between features generated from different modalities may lead to an incorrect mimic of the noisy features and inaccurate 3D geometry representation. Therefore, we propose a geometry compensation module to address the geometry misalignment and handle the modal difference.

\noindent\textbf{Geometry Compensation Module (GCM)}\label{sec:compensation}
A crucial process in the multi-view camera-based detection method is the view projection operation, which transforms camera-view (UV) features into the BEV space. Inaccurate geometry inference in this process leads to geometry misalignment between features learned from images and LiDAR, exacerbating the modality gap. Therefore, we propose to conduct geometry compensation before and after the view projection in the simulated LiDAR branch to learn more accurate geometry features in both UV and BEV space. 

Deformable Convolutions and Deformable Attentions are known to be effective in enabling neural networks to model spatial transformations and account for geometric deformations or misalignments \cite{dai2017deformable,zhu2020deformable}. Therefore, we adopt deformable self-attention layers to construct GCM, as shown in Figure \ref{fig:GCMillustration}.
For geometry compensation in the UV space, we first generate a uniform grid of points $Q_{uv}$ as query points for each 2D camera feature $F^{S}_{C_{uv}}$. Then, we learn offsets based on each point $q_{(u,v)}\in Q_{uv}$ to generate a set of most related points $\mathcal{P}_{uv}$ around it. These learned points $\mathcal{P}_{uv}$ are taken as reference points and keys used to sample the value features from the 2D camera features $F^{S}_{C_{uv}}$. With the optimization signals gradually improving attentive locations, the module facilitates the model to compensate for geometric transformations in the x-y plane. We apply standard multi-head attention, learning individual offsets for each head, which captures abundant information and improves feature representations for subsequent context learning, depth estimation, and 3D geometry inference. Similarly, we employ a BEV geometry compensation module after transforming the camera-view features to BEV features $F^{S}_{C_{uv-bev}}$, which is responsible for correcting the key feature locations in the x-z plane. By doing so, the geometry compensation in the two complementary 2D views can comprehensively improve the feature representation. Overall, the view projection with GCM used in the simulated LiDAR branch is formulated here, with reference to Eq.~\eqref{eq:bevfeature}:
\begin{equation}
\begin{split}
    F^{S}_{L_{bev}}&= Proj_L(F^{S}_{C_{uv}}) \\
    &= GC_{bev}(\rho(\psi(GC_{uv}(F^{S}_{C_{uv}})) \times \phi(GC_{uv}(F^{S}_{C_{uv}})))),
\end{split}
\label{eq:gcm}
\end{equation} 
where $GC_{uv}(F^{S}_{C_{uv}})=\rm DeformAttn(Q_{uv}, \mathcal{P}_{uv}, \mathnormal{F^{S}_{C_{uv}})}$
and $GC_{bev}(F^{S}_{C_{uv-bev}})=\rm DeformAttn(Q_{bev}, \mathcal{P}_{bev}, \mathnormal{F^{S}_{C_{uv-bev}}})$, denoting the UV Geometry Compensation and BEV Geometry Compensation, respectively. $Q_{bev}$ and $\mathcal{P}_{bev}$ are query and reference points generated for BEV features $F^{S}_{C_{uv-bev}}$.

To get the final simulated LiDAR feature for distillation, we also implement a simple yet effective object-aware mask $\mathcal M$ to select the most informative features at the end of GCM. We generate masks in the BEV space from the ground truth center points and bounding boxes using a heatmap-like approach like BEVDisitll \cite{chen2023bevdistill}. Therefore, the CMD loss is formulated as:
\begin{equation}
    \begin{split}
        \mathcal L_{CMD} &=  {\rm MSE}(\mathcal M \odot F^{T}_{L_{bev}}, \mathcal M \odot F^{S}_{L_{bev}}),
    \end{split}
\end{equation}where $\odot$ is Hadamard product.
The object-aware mask is a technique we utilize together with GCM to improve the ability to overcome the cross-modal gap in CMD and we refrain from attributing it as our original contribution.

\begin{table*}[!htp]
\footnotesize
\setlength{\tabcolsep}{0.006\linewidth}
\resizebox{\textwidth}{!}{
\centering
\begin{tabular}[t]{c|c|c|c|c|c|c|c|c|c|c}
\toprule
Methods &Modality  & Backbone& Image Size & mAP$\uparrow$ &NDS$\uparrow$ &mATE$\downarrow$ &mASE$\downarrow$ &mAOE$\downarrow$ &mAVE$\downarrow$ &mAAE$\downarrow$ \\
\midrule
BEVFusion \cite{liang2022bevfusion}&LC &VoxelNet SwinT &$448\times 800$   & 67.9 &71.0 & -& -& - & -&-  \\

BEVFusion \cite{liu2022bevfusion}&LC &VoxelNet SwinT&$256\times 704$  & 68.5 &71.4 &28.6  &25.3&30.0&25.4&18.6\\
  
FCOS3D \cite{wang2021fcos3d}&C &R101 &$900\times 1600$&  29.5 &37.2 & 80.6&26.8& 51.1&113.1 &17.0 
  \\
 
 
  

 BEVDet \cite{huang2021bevdet} &C& R50 &$256\times 704$ &  29.8 & 37.9 & 72.5 &27.9 &58.9 &86.0 &24.5\\
  
   PETR \cite{liu2022petr} &C& R50 &$384\times 1056$&   31.3 &38.1 &76.8 &27.8 &56.4 &92.3 &22.5 \\
 
DETR3D \cite{huang2022bevdet4d} &C&R101 &$900\times 1600$&   34.9 &43.4 & 71.6&26.8& 37.9&84.2 &20.0 
 \\
\midrule
 \midrule
Set2Set \cite{li2022unifying}&C*& R50 &$900\times 1600$ &33.1 &41.0&-
 &-&-&-&-\\
  MonoDistill \cite{chong2021monodistill}&C*& R50 &$900\times 1600$ &36.4 &42.9&-
 &-&-&-&-\\
 UVTR \cite{li2022unifying}&C*& R50 &$900\times 1600$ &36.2 &43.1&-
 &-&-&-&-\\

TiG-BEV \cite{huang2022tig}&C*& R50 &$256\times 704$ & 33.1 &41.1&67.8
 &27.1&58.9&78.4&21.8\\
 UniDistill \cite{zhou2023unidistill} &C*& R50 &$256\times 704$ &26.5&37.8&-&-&-&-&-\\
 BEVDistill \cite{chen2023bevdistill} &C*& SwinT &$256\times 704$ &36.3  &43.6&64.2
 &27.4&57.6&87.8&28.2\\
 BEVFusion-C \cite{liu2022bevfusion} &C& SwinT &$256\times 704$ & 35.6 &41.2&66.8
 &27.3&56.1&89.6&25.9\\
 
 SimDistill &C*& SwinT &$256\times 704$ & 40.4 &45.3&52.6
 &27.5&60.7&80.5&27.3\\
 \bottomrule
\end{tabular}}
\caption{Quantitative comparisons on the nuScenes validation Set. L and C in the second column denote the input modality, \ie, LiDAR and camera, while C* means using LiDAR for knowledge distillation during training. 
}
\label{tab:maintable}
\end{table*}
\subsubsection{Multi-modal fusion Distillation}\label{sec:MMD}
In light of the aligned architecture and workflow with the teacher model, the student model also produces multi-modal fusion features as well as detection predictions. To make the fused feature and predictions consistent with those in the teacher model, we devise multi-modal distillation in both feature level (MMD-F) and prediction level (MMD-P). Owing to the proximity of the fusion module and the detection head, MMD-F is expected to distill highly useful multi-modal knowledge that directly contributes to the detection. It is implemented by aligning the fusion feature of the teacher model and the simulated fusion feature of the student model:
\begin{equation}
     \begin{split}
         \mathcal L_{MMD-F} &= {\rm MSE}(F^T_{U_{bev}}, F^S_{U_{bev}}). \\
     \end{split}
 \end{equation}
After the fusion module, the fused feature in the student model is fed into the detector to output the detection results in the same way as the teacher model. Thus, we also employ MMD-P by taking the predictions from the teacher model as soft labels. We adopt the quality-aware prediction distillation loss $L_{MMD-P}$ \cite{hong2022cross}, which consists of the classification loss $\mathcal L_{cls}$ for object categories and the regression loss $\mathcal L_{reg}$ for 3D bounding boxes:
\begin{equation}
\begin{split}
    \mathcal L_{MMD-P} &= \mathcal L_{reg} + \mathcal L_{cls},\\
        &={\rm SmoothL1}(P^T_B,P^S_B) \cdot s + {\rm QFL}(P^T_C,P^S_C) \cdot s,
\end{split}
\end{equation}
where $P^T_B$ and $P^T_C$ (resp. $P^S_B$ and $P^S_C$) denote the predicted bounding boxes and categories by the teacher model (resp. the student model). $\rm QFL(\cdot)$ denotes the quality focal loss~\cite{li2020generalized}. $s$ is a quality score used as the loss weight, obtained by measuring the IoU between the predictions and the ground truth to determine the confidence of the soft label.

\noindent\textbf{Discussion}
It is noteworthy that previous methods have not explored multi-modal fusion distillation due to the absence of a dedicated multi-modal architecture in the student model for aligning fusion features or predictions. Instead, these methods distill information solely by aligning the teacher model's fusion features or predictions to a single-modal student counterpart, which leads to subpar performance due to the modality gap. Furthermore, no studies have investigated the impact of comprehensive multi-modal distillation, including intra-modal, cross-modal, and multi-modal fusion distillation, simultaneously. Our SimDistill makes progress by effectively performing multi-modal fusion distillation through its simulated multi-modal architecture. This complements intra-modal and cross-modal distillation (Sec.~4.3), resulting in improved performance.

\subsection{Training Objective}
\label{sec:loss}
Apart from the above distillation losses, the student model is also optimized by the common loss of 3D object detection task $\mathcal{L}_{det}$. The overall training objective $\mathcal{L}$ is defined as:
\begin{equation}
    \mathcal{L} = \mathcal L_{IMD} + \mathcal L_{CMD}+ \mathcal L_{MMD-F}+ \mathcal L_{MMD-P} + \mathcal{L}_{det}.
\end{equation}

\section{Experiment}
\subsection{Experiment Setting}
\noindent\textbf{Datasets and Evaluation Metrics} We follow the common practice \cite{huang2021bevdet,liu2022bevfusion,liang2022bevfusion,li2022bevdepth,chen2023bevdistill} to evaluate our method on the most challenging benchmark, \ie, nuScenes \cite{caesar2020nuscenes}. It comprises 700 scenes for training, 150 scenes for validation, and 150 scenes for testing. Each scene includes panoramic LiDAR data and surrounding camera images, which are synchronized to provide convenience for multi-modal-based research. The dataset comprises a total of 23 object categories, and 10 popular classes are considered for computing the final metrics. To align with the official evaluation, we adopt mean Average Precision (\textbf{mAP}) and nuScenes detection score (\textbf{NDS}) as the main metrics with other 5 metrics for reference.

\noindent\textbf{Implementation Details} 
Our method is implemented with PyTorch using 8 NVIDIA A100 (40G Memory), based on the MMDetection3D codebase \cite{mmdet3d2020}. We adopt BEVFusion \cite{liu2022bevfusion} as the default teacher model, which takes images with a size of $256\times704$ and LiDAR point cloud with a voxel size of (0.075m, 0.075m, 0.2m) as input and uses VoxelNet \cite{zhou2018voxelnet} and Swin-T \cite{liu2021swin} as backbones for the two modalities, respectively. During distillation, we utilize the official BEVFusion checkpoint, freeze the teacher model, and train the student model for 20 epochs with batch size 24. The backbone and input resolution are kept the same as BEVFusion-C in both our SimDisitll and our competitor BEVDistill. More implementation details, ablation analysis, and visualizations can be found in Appendices.

\subsection{Main Results}
We compare our SimDistill with state-of-the-art methods on the nuScenes validation set and present the results in Table~\ref{tab:maintable}. We group the methods according to the input modality and present the knowledge distillation-based methods in the bottom part (except for baseline BEVFusion-C) for straightforward comparisons. From the table, we can see that fusion-based methods usually possess a stronger perception ability and achieve better performance. However, the high cost of LiDAR may restrict their practical usage. Compared with the baseline BEVFusion-C, SimDistill boosts the performance significantly by 4.8\% mAP and 4.1\% NDS, clearly validating the effectiveness of the proposed distillation method. 
Compared with the concurrent distillation methods BEVDistill and UniDistill,
our SimDistill achieves much better performance under the same setting.

\subsection{Ablation Studies}

\begin{table}[t]
\footnotesize
\setlength{\tabcolsep}{0.02\linewidth}
\centering
\resizebox{0.475\textwidth}{!}{
\begin{tabular}[t]{c|c|c|c|c|c}
\toprule
 & Teacher & Student & Distillation & mAP$\uparrow$ & NDS$\uparrow$ \\
\midrule
a&BEVFusion&BEVFusion-C&MMD-F &35.94&41.75 \\
b&BEVFusion&SimDistill &MMD-F&38.34&44.15 \\
c&BEVFusion-L&BEVFusion-C &CMD-v&35.88&42.87 \\
d&BEVFusion-L&SimDistill &CMD-v&36.80&42.79 \\
 \bottomrule
\end{tabular}}
\caption{Ablation study of the model architecture. CMD-v is the vanilla version of CMD without using GCM here.}
\label{tab:architecture}
\end{table}

\paragraph{Why Choose Multi-modal Architectures?}~\\
To demonstrate the superiority of the proposed simulated multi-modal structure,
we replace the multi-modal teacher BEVFusion and the simulated multi-modal student SimDistill with their single-modal counterpart BEVFusion-L (\ie, the LiDAR branch of BEVFusion) and BEVFusion-C, respectively. The results are presented in Table \ref{tab:architecture}. We first investigate the influence of using a simulated multi-modal student. In models (a) and (b), we adopt the multi-modal teacher (BEVFusion) but distill the fusion feature to different student architectures. The experiment results show that the simulated multi-modal student (b) outperforms the single-modal one (a) with a clear gain of 2.4 in both mAP and NDS. We then change the teacher to a single-modal one (BEVFusion-L) to verify the performance of the student. Although directly learning from a cross-modal teacher adversely affects performance due to the modality gap, the multi-modal student (d) still achieves better performance in mAP and comparable results in NDS compared with the single-modal student (c). The two groups of comparisons validate the superiority of using a multi-modal student. Besides, the experiments of (b) and (d) both directly distill the learned feature from the teacher model to the student, which validates the importance of using a multi-modal teacher, \ie, with a gain of 1.54\% mAP and 1.36\% NDS. In summary, it is crucial to employ multi-modal architectures for both teacher and student models to enhance knowledge transfer and achieve better performance. In addition, employing the proposed simulated multi-modal student model maintains the advantage of cost-effective camera-only deployment.

\paragraph{How Simulated Multi-modal Distillation Works?}~\\
To investigate the impact of distillation options, we perform ablation studies and summarize the results in Table \ref{tab:ablation}. Model (a) denotes the baseline model with the proposed simulated multi-modal architecture without any knowledge distillation. We present the gains over Model (a) in the column mAP and NDS. 
As shown in (b), (c), (e), and (f), employing IMD, vanilla CMD, MMD-F, and MMD-P on the baseline model leads to 1.09\%, 1.43\%, 2.63\%, and 1.02\% absolute gains in mAP, respectively, where MMD-F brings the largest gain owing to the rich multi-modality knowledge contained in the fusion features. Interestingly, while the simulated LiDAR branch should possess more accurate 3D geometry than the camera branch, IMD (b) produces a slightly larger gain than vanilla CMD (c). We attribute it to the modality gap between the real LiDAR features of the teacher and the simulated ones of the student. 

\begin{table}[t]
\footnotesize
\setlength{\tabcolsep}{0.015\linewidth}
\centering
\resizebox{0.475\textwidth}{!}{
\begin{tabular}[t]{c|c|c|c|c|c|c|c}
\toprule
& \multirow{2}{*}{IMD}& \multicolumn{2}{c|}{CMD}   & \multicolumn{2}{c|}{MMD}&\multirow{2}{*}{mAP$\uparrow$}&\multirow{2}{*}{NDS$\uparrow$ }\\
\cline{3-6}
&  &vanilla&GCM&   -F&-P&&\\
\hline
a    && &&&& 35.71 (-)  &41.97 (-) \\
\midrule
b&$\checkmark$&&   &  & & 37.14 (+1.43) &42.67 (+0.70)\\
\midrule
c &&$\checkmark$   &&   &&  36.80 (+1.09) &42.79 (+0.82) \\

d &&$\checkmark$  & $\checkmark$ &&  &39.83 (+4.12)  &44.79 (+2.82)\\
\midrule
e&&&  &  $\checkmark$ &  & 38.34 (+2.63) &44.15 (+2.18) \\
f&&&  &   &$\checkmark$& 36.73 (+1.02) &42.52 (+0.55)\\
\midrule
g &$\checkmark$ & $\checkmark$ &$\checkmark$ &$\checkmark$   &$\checkmark$ & 40.40 (+4.69) &45.31 (+3.34) \\
 \bottomrule
\end{tabular}}
\caption{Ablation study of different distillation options. 
}
\label{tab:ablation}
\end{table}
\begin{table}[t]
\footnotesize
\centering
\resizebox{0.475\textwidth}{!}{
\begin{tabular}[t]{c|cc|cc}
\toprule
Methods&FPS &GFlops &mAP&NDS\\
\midrule
BEVDet \cite{huang2021bevdet}&15.6&215.3&31.2&39.2\\
BEVFormer \cite{li2022bevformer}&2.4&1303.5&37.5&44.8 \\
BEVDistill\cite{chen2023bevdistill}&3.7&608.8&36.3&43.6 \\
BEVFusion-C \cite{liu2022bevfusion}&13.4 &165.1&35.6&41.2\\
SimDistill&11.1&219.1&40.4&45.3\\
 \bottomrule
\end{tabular}}
\caption{Comparison of model efficiency.}
\label{tab:efficiency}
\end{table}

After using the proposed GCM (\ie, Model (d)), we can see that it helps CMD achieve a gain of 4.12\% mAP and 2.82\% NDS over the baseline in (a), validating the effectiveness of GCM in overcoming the side effect of the modality gap during distillation. After incorporating all the components including simulated multi-modal architecture and all the distillation techniques, we get our SimDistill model in (g), which delivers the best performance of 40.40 mAP and 45.31 NDS, meanwhile achieving an improvement of 4.8\% mAP and 4.1\% NDS over the baseline model BEVFusion-C.

\subsection{Model Efficiency}
 We compare the model efficiency with other representative methods in Table~\ref{tab:efficiency}. Our method achieves an inference speed of 11.1 FPS on a single GPU, running much faster than BEVDistill and BEVFormer. It is comparable to BEVFusion-C but a bit slower than BEVDet, mainly due to the additional simulated LiDAR branch in the architecture. Nevertheless, SimDistill significantly outperforms other methods in terms of mAP and NDS. 


\section{Conclusion}
In this paper, we propose a novel simulated multi-modal distillation method named SimDistill for multi-view BEV 3D object detection by carefully investigating the architecture design and effective distillation techniques. We identify the importance of the multi-modal architecture for multi-modal knowledge distillation and devise a simulated multi-modal student model accordingly. Built upon it, we develop a novel simulated multi-modal distillation scheme that supports intra-modal, cross-modal, and multi-modal fusion knowledge distillation simultaneously. Experiments on the challenging nuScenes benchmark have validated the above findings and the superiority of the proposed distillation methods over state-of-the-art approaches. We believe SimDistill is compatible with other multi-modal teacher and diverse student models, which could lead to enhanced performance and remains a subject for future investigation.
\section*{SimDistill: Simulated Multi-modal Distillation for BEV 3D Object Detection (Appendices)}
\section{Network Architecture}
\subsection{Details of the Model Configuration}
To clearly illustrate the structure of our method, we supplement a detailed description of the network architecture in Table \ref{tab:001}. 

The student model of SimDistill comprises the following components: a 2D encoder, two parallel view projections in the camera and LiDAR branch, a fusion module, and a detection head. The 2D encoder is composed of a Swin Transformer-tiny backbone \cite{liu2021swin} and an FPN (Feature Pyramid Network) neck \cite{lin2017feature}. After extracting features with the 2D encoder, the pipeline is split into two branches, namely the camera branch and the simulated LiDAR branch, to conduct the view projection and obtain the corresponding BEV (Bird's-Eye-View) features.

\begin{table}[h]
\scriptsize
    \begin{center}
    \begin{tabular}{|c|c|c|c|c|}
        \hline
        \multicolumn{4}{|c|}{\bf{2D Encoder}}\\
         \hline
         \bf{name} &\bf{input channel} &\bf{operation} &\bf{output channel}	\\
         \hline
      encoder backbone&[3]	&Swin-T 	& [192, 384, 768]		 	\\
        \hline
      encoder neck& [192, 384, 768]	&FPN 			&[256]		 \\
         
         \hline
        \hline
         \multicolumn{4}{|c|}{\bf{View Projection in Camera branch (parallel)}}\\
         
         \hline
         \bf{name} &\bf{input channel}&\bf{operation}&\bf{output channel}	\\
         \hline
        depthNet&[256]	&Conv2d	&[198]		\\
         \hline
         lift&[198]	& $\times$	&[80*118]		\\
         \hline
         bevPool&[80*118]	& flatten	&[80]		\\
          \hline
         downsample&[80]	& Conv2d	&[80]		\\
         \hline
        \hline
        \multicolumn{4}{|c|}{\bf{View Projection in Simulated LiDAR branch (parallel)}}\\
     \hline
         \bf{name}  &\bf{input channel}&\bf{operation}&\bf{output channel}	\\
         \hline
         UV GCM &[256]	&DeformAttn	&[256]		\\
         \hline
        depthNet&[256]	&Conv2d	&[198]		\\
         \hline
         lift&[198]	& $\times$	&[80*118]		\\
         \hline
         bevPool&[80*118]	& flatten	&[80]		\\
          \hline
         downsample&[80]	& Conv2d	&[80]		\\
         \hline
         BEV GCM&[80]	& DeformAttn	&[80]		\\
         \hline
         \hline
        \multicolumn{4}{|c|}{\bf{Fusion Module}}\\
         \hline
         \bf{name}  &\bf{input channel}&\bf{operation}&\bf{output channel}	\\
         \hline
        fuser	&[160]	&Conv2d	&[256]			\\
       \hline
        decoder backbone	&[256]	&ResNet Block	&[512]			\\
        \hline
        decoder neck	&[512]	&FPN	&[256]			\\
        \hline
        \hline
        \multicolumn{4}{|c|}{\bf{Head}}\\
         \hline
          \bf{name}  &\bf{input channel}&\bf{operation}&\bf{output channel}	\\
         \hline
        head	&[256]	&CenterHead	&-			\\
       \hline
\end{tabular}
\end{center}
    \caption{The architecture of the student model of SimDistill. We list all the important modules and operations in the model.}
    \label{tab:001}
\end{table}{}
In the view projection module of the camera branch, the extracted features are passed through a DepthNet comprising 3 Conv2d layers to obtain the depth probabilities. The default range of depth prediction is $[1, 60]$ meters, with intervals of $0.5$ meters, resulting in an output channel of $198$ ($80$ channels for the context feature and $(60-1) / 0.5 =118$ channels for the depth feature).

The 2D features are then lifted to 3D space using the LSS method \cite{philion2020lift}. Each 2D feature pixel is scattered into $D$ discrete points along the camera ray by rescaling the context feature with their corresponding depth probabilities. The resulting feature point cloud is then aggregated and flattened into BEV grids using a BEV pooling operation. After downsampling, the final BEV feature map has a size of $180 \times 180$ with $80$ channels.

The simulated LiDAR branch is parallel and similar to the camera branch but with two additional geometry compensation modules (GCMs), \ie, UV GCM and BEV GCM, which are formed of two multi-head deformable self-attention layers. The UV GCM is imposed on the camera-view 2D features before the view projection to refine the geometry information for obtaining more informative and precise BEV features. Similarly, the BEV GCM is applied to the BEV feature to further improve and adjust the geometry information to better align with the BEV features from the LiDAR branch of the teacher model.

After obtaining the BEV features of the two branches, the features are fused using a convolutional fusion module, following the teacher model BEVFusion \cite{liu2022bevfusion}. Last, the CenterHead \cite{yin2021center} is leveraged to generate the final detection results.
\begin{figure*}[t]
\begin{center}
  \includegraphics[width=0.935\linewidth]{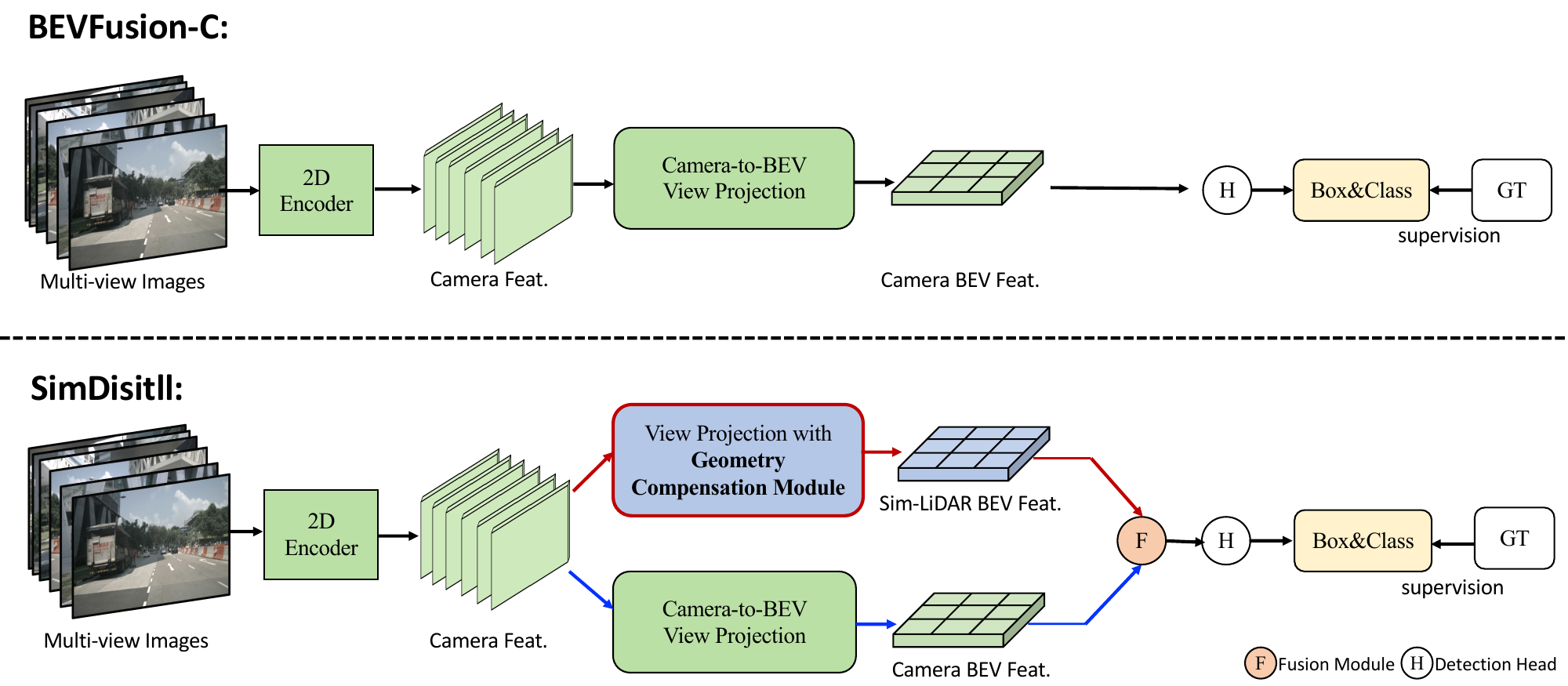}
\end{center}
  \caption{Comparison of the overall pipeline of BEVFusion-C and our SimDisitll (student model). Both models take multi-view images as input while our method extends into two branches to additionally learn simulated LiDAR features and conduct multi-modal knowledge distillation.}
\label{fig:modelcomparison}
\end{figure*}
\subsection{Comparison of SimDistill and BEVFusion-C}
To clearly demonstrate the difference in model architecture between SimDisill and our baseline method BEVFusion-C, we illustrate the network pipelines in Figure \ref{fig:modelcomparison}. Both models take multi-view images as input while our method extends into two branches to additionally learn simulated LiDAR features and conduct multi-modal knowledge distillation. Our SimDistill boosts the baseline method BEVFUsion-C with a gain of 4.8\% mAP and 4.1\% NDS. Notably, our SimDistill achieves this enhancement without significant alterations to the original model architecture or imposing excessive computational overhead.

\begin{table*}[htbp]
\footnotesize
\centering
\resizebox{\textwidth}{!}{
\begin{tabular}[t]{c|c|c|c|c|c|c|c|c}
\toprule
Methods &Modality &Frames & Backbone& Image Size & mAP$\uparrow$ &NDS$\uparrow$ &mATE$\downarrow$ &mASE$\downarrow$ \\
\midrule
CenterPoint\cite{yin2021center}&L & 1&VoxelNet &-  & 59.6 &66.8 & - & -\\
BEVFusion \cite{liang2022bevfusion}&LC & 1&VoxelNet SwinT &$448\times 800$   & 67.9 &71.0 & -& - \\

DeepInteraction \cite{yang2022deepinteraction}&LC &1&VoxelNet R50 &$450\times 800$ & 69.9 &72.6 & -& -\\
BEVFusion \cite{liu2022bevfusion}&LC& 1 &VoxelNet SwinT&$256\times 704$  & 68.5 &71.4 &28.6  &25.3\\
  \midrule
  \midrule
  
FCOS3D \cite{wang2021fcos3d}&C & 1&R101 &$900\times 1600$&  29.5 &37.2 & 80.6&26.8\\

 BEVFormer \cite{li2022bevformer} &C& 3&R101 &$900\times 1600$&  41.6 &51.7 & 67.3&27.4\\
 
  
 PolarFormer \cite{chen2022polar} &C&2&R101 &$900\times 1600$&  43.2
 & 52.8&64.8&27.0\\

 BEVDet \cite{huang2021bevdet} &C& 1& R50 &$256\times 704$ &  29.8 & 37.9 & 72.5 &27.9\\ 
  
   PETR \cite{liu2022petr} &C&1& R50 &$384\times 1056$&   31.3 &38.1 &76.8 &27.8 \\
 
DETR3D \cite{huang2022bevdet4d} &C& 2&R101 &$900\times 1600$&   34.9 &43.4 & 71.6&26.8\\
 CrossDTR \cite{tseng2023crossdtr}&C& 1&R101 &$900\times 1600$&   37.0 &42.6 & 77.3&26.9\\
 \hline
 BEVDepth \cite{li2022bevdepth} &C* &2& R50 &$256\times 704$  &35.1 &47.5&63.9
 &26.7\\
 
STS \cite{wang2022sts}&C*&2 & R50 &$256\times 704$  &37.7 &48.9&60.1
 &27.5\\

BEVStereo \cite{li2023bevstereo} &C*&2& R50 &$256\times 704$ &37.2 &50.0&59.8
 &27.0\\
 SOLOFusion \cite{park2022time} &C*&16& R50 &$256\times 704$ & 42.7 &53.4&56.7
 &27.4\\
\midrule
 \midrule
Set2Set \cite{li2022unifying}&C*&1& R50 &$900\times 1600$ &33.1 &41.0&-
 &-\\
  MonoDistill \cite{chong2021monodistill}&C*&1& R50 &$900\times 1600$ &36.4 &42.9&-
 &-\\
 UVTR \cite{li2022unifying}&C*&1& R50 &$900\times 1600$ &36.2 &43.1&-
 &-\\
 BEVDistill \cite{chen2023bevdistill}&C*&1& R50 &$900\times 1600$ &38.6 &45.7&69.3
 &26.4\\

TiG-BEV \cite{huang2022tig}&C*&1& R50 &$256\times 704$ & 33.1 &41.1&67.8
 &27.1\\
 UniDistill \cite{zhou2023unidistill} &C*&1& R50 &$256\times 704$ &26.5&37.8&-&-\\
 BEVFusion-C \cite{liu2022bevfusion} &C&1& SwinT &$256\times 704$ & 35.5 &41.2&66.8
 &27.3\\
 BEVDistill \cite{chen2023bevdistill} &C*&1& SwinT &$256\times 704$ &36.3  &43.6&64.2
 &27.4\\
 Ours &C*&1& SwinT &$256\times 704$ & \textbf{40.4} &{45.3}&52.6
 &27.5\\
  Ours &C*&1& ViTAEv2-S &$256\times 704$ & {40.1} &\textbf{46.3}&51.1
 &27.4\\
 \bottomrule
\end{tabular}}
\caption{Quantitative comparisons on the nuScenes validation Set. L and C in the second column denote the input modality, \ie, LiDAR and camera, while C* means using LiDAR for depth supervision or knowledge distillation during training. 
Frames denotes the number of temporal frames used during inference.}
\label{tab:maintablesup}
\end{table*}
\subsection{Implementation Details}
We take the well-trained BEVFusion model \cite{liu2022bevfusion} as our teacher model and freeze it during the distillation. 
When training the student, we simply sum up all the loss items without heavily adjusting the weight parameters.
AdamW \cite{loshchilov2017decoupled} is used as the optimizer during training with an initial learning rate of 2e-4 and a cyclic policy. The same data augmentation used in the camera branch of the BEVFusion model is used in the student model. During the evaluation, only the student model is used with one single frame as input and without any test-time augmentation.
\section{Additional Quantitative Experiments}
Due to the space limit, we only list related methods under similar settings to ours in the main submission. Here, we list more recent methods with different settings in Table \ref{tab:maintablesup} to conduct a more comprehensive comparison. For these state-of-the-art methods, there are diverse settings regarding backbones, input resolutions, and the number of temporal frames leveraged during evaluation. Generally, methods with heavier backbone, larger input image size, and more input frames will achieve better performance. Due to the difficulty of comparing all the methods under the same setting, we mainly compare SimDistill with those distillation-based under similar settings, as shown in Table \ref{tab:maintablesup} and Table \ref{tab:backbone}.
\subsection{Comparison Methods}
Although there are several related works including Monodistill \cite{chong2021monodistill}, BEVDistill \cite{chen2023bevdistill}, and Unidistill \cite{zhou2023unidistill}, focusing on knowledge distillation on the 3D object detection task, SimDistill differs from these methods in both architecture and distillation scheme. Monodistill is dedicated to cross-modal distillation between a LiDAR-based teacher and a camera-based student in the monocular image view instead of the multi-camera Bird's-eye view. And the student model in Monodistill learns affinity maps instead of features to address the modality gap, whereas our method incorporates a dedicated branch for learning LiDAR features and introduces GCM to bridge the modality gap. BEVDistill also conducts knowledge transfer in BEV space but it is a cross-modal distillation method as well, missing the opportunity of leveraging a stronger fusion-based teacher model.
The most relevant work to ours is Unidistill. Although it similarly incorporates a multi-modal teacher, it functions as a general knowledge distillation framework that accommodates both single-to-single and fusion-to-single cross-modal distillation scenarios. However, it overlooks the concern of architectural disparities between the teacher and student models and lacks the capacity to thoroughly execute multi-modal distillation and bridge the cross-modal disparity. Experiment results in Table \ref{tab:maintablesup} show the superiority of our methods over these competitors under the same setting. 

\subsection{Evaluation Metrics}
We employ the official evaluation metrics provided by nuScenes, encompassing two main metrics mean Average Precision (mAP) and nuScenes detection score (NDS), and five additional metrics which are mean Average Translation Error (mATE), mean Average Scale Error (mASE), mean Average Orientation Error (mAOE), mean Average Velocity Error (mAVE) and mean Average Attribute Error (mAAE). The mAP evaluates both recall and precision pertaining to predicted bounding boxes. NDS represents an amalgamation of the other five metrics, offering a comprehensive assessment of detection capability. The measurement unit for these results is expressed as a percentage (``\%").
\subsection{The Impact of Backbone}
To further evaluate our approach, we conducted additional quantitative experiments using different backbones, including ResNet-50 \cite{he2016deep} and ViTAEv2-S \cite{zhang2023vitaev2}, as shown in Table \ref{tab:backbone}. Swin-T and ViTAEv2-S have similar parameter quantities as ResNet-50, but they exhibit stronger representation abilities, resulting in superior performance. We also compare with other distillation-based 3D object detection methods in Table \ref{tab:backbone}, demonstrating our superiority under different settings.
\begin{table}[t]
\footnotesize
\resizebox{0.475\textwidth}{!}{
\centering
\begin{tabular}[t]{c|c|c|c|c}

  \toprule
Methods &Mod  & Backbone & mAP$\uparrow$ &NDS$\uparrow$ \\
\midrule
 BEVFusion-C \cite{liu2022bevfusion} &C& R50  & 31.6 &39.3\\
 TiG-BEV \cite{huang2022tig}&C*& R50  &33.1&41.1\\
 UniDistill \cite{zhou2023unidistill}&C*& R50&26.5&37.8\\
 SimDisitll &C*& R50  &37.3  &43.8\\
 \midrule
BEVFusion-C \cite{liu2022bevfusion} &C& Swin-T  & 35.5 &41.2\\
BEVDistill \cite{chen2023bevdistill}&C*& Swin-T  &36.3&43.6\\
 SimDistill &C*& Swin-T  & \textbf{40.4} &\underline{45.3}\\
 \midrule
  SimDisitll &C*& ViTAEv2-S  &\underline{40.1} &\textbf{46.3}\\
 \bottomrule
\end{tabular}}
\caption{Quantitative comparisons on the nuScenes validation Set. C in the Mod column means the models' input modality is camera-only, while C* means imposing LiDAR for knowledge distillation during training. Swin-T and ViTAEv2-S are backbone networks with similar quantities of parameters with ResNet-50 (R50). The input image size for all these models is $256 \times 704$ and only a single frame is used for inference.}
\label{tab:backbone}
\end{table}
\begin{table}[t]
\footnotesize
\centering
\resizebox{0.475\textwidth}{!}{
\begin{tabular}[t]{|c|c|c|c|}
\hline

  Camera & Sim-LiDAR& mAP$\uparrow$ & NDS$\uparrow$ \\
\hline
\multicolumn{2}{|c|}{BEVFusion-C baseline}&35.6&41.2\\
\hline
BEVFusion-C&BEVFusion-C &$40.4_{+4.8}$&$45.3_{+4.1}$ \\
\hline
\multicolumn{2}{|c|}{{BEVFormer baseline}}&{37.5}&{44.8}\\
\hline
BEVFormer &BEVFusion-C&$44.2_{+6.7}$&$45.2_{+0.4}$\\
 \hline
\end{tabular}}
\caption{We replace the original BEVFusion-C with BEVformer in the camera branch of the student model to demonstrate the flexibility of SimDistill.}
\label{tab:architectureSup}
\end{table}
\subsection{Flexibility of SimDisitll}
To demonstrate the flexibility and adaptability of our SimDistill, we also conduct an experiment by changing  the student model. In the default SimDisitll, the student model is designed based on the camera branch of BEVFusion \cite{liu2022bevfusion}, \ie, BEVFusion-C. As shown in Table \ref{tab:architectureSup}, we replace BEVFusion-C with BEVFormer \cite{li2022bevformer} in the camera branch of the student model. According to the experiment results, the adapted version (b) further improves the default version (a) of SimDistill by a gain of  4.2 \% mAP. This indicated the potential for further expansion and improvement of SimDistill. In future work, we will be dedicated to adapting our SimDistill to more diverse and advanced teacher and student models to further explore upper-bound performance.
\subsection{More Ablation Study}
\subsubsection{2.3.1 Ablation Study for the complementary effect of distillation options}
Except for the ablation study in the main submission, we report more experiment results to analyze the effect of diverse components in our method and their complementary effect, shown in Table \ref{tab:ablationSup}. According to the results of Modal (b), (c) and (d), IMD and CMD are demonstrated to be complementary and can be utilized together to further improve the performance. For the multi-modal fusion distillation (MMD) part, although solely using MMD-F (f) and MMD-P (g) can both improve the performance, directly combining together (h) without any single-modal feature distillation IMD and CMD will cause a slight performance decrease, which may be caused by the confusion for the two branch features learning due to the conflicting signal from MMD-F and MMD-P. This phenomenon is addressed in Model (i) with the additional loss items IMD and CMD in two branches, bringing the largest performance gains without the proposed GCM, \ie 3.18\% and 2.45\% in mAP and NDS. Therefore, to conduct multi-modal fusion distillation, it must be utilized together with single-modal distillation IMD and CMD simultaneously, demonstrating the complementary effect of IMD, CMD and MMD. Besides, by adding GCM in the final Model (j), the method achieves gains of 4.69\% mAP and 3.34\% NDS, demonstrating the complementary effectiveness of GCM with the whole distillation scheme. 
\begin{table}[t]
\footnotesize
\resizebox{0.475\textwidth}{!}{
\centering
\begin{tabular}[t]{c|c|c|c|c|c|c|c}
\toprule
& \multirow{2}{*}{IMD}& \multicolumn{2}{c|}{CMD}   & \multicolumn{2}{c|}{MMD}&\multirow{2}{*}{mAP$\uparrow$}&\multirow{2}{*}{NDS$\uparrow$ }\\
\cline{3-6}
&  &vanilla&GCM&   -F&-P&&\\
\hline
a    && &&&& 35.71 (-)  &41.97 (-) \\
\midrule
b&$\checkmark$&&   &  & & 37.14 (+1.43) &42.67 (+0.70)\\
\midrule
c &&$\checkmark$   &&   &&  36.80 (+1.09) &42.79 (+0.82) \\
d &$\checkmark$&$\checkmark$   &   &&&  37.30 (+1.59) &43.0 (+1.03) \\
e &&$\checkmark$  & $\checkmark$ &&  &39.83 (+4.12)  &44.79 (+2.82)\\
\midrule
f&&&  &  $\checkmark$ &  & 38.34 (+2.63) &44.15 (+2.18) \\
g&&&  &   &$\checkmark$& 36.73 (+1.02) &42.52 (+0.55)\\
h&&&  & $\checkmark$  &$\checkmark$& 38.17 (+2.46) &43.20 (+1.23)\\
j&$\checkmark$&$\checkmark$&  & $\checkmark$ &$\checkmark$& 38.89 (+3.18) &44.42 (+2.45)\\
\midrule
k &$\checkmark$ & $\checkmark$ &$\checkmark$ &$\checkmark$   &$\checkmark$ & 40.40 (+4.69) &45.31 (+3.34) \\
 \bottomrule
\end{tabular}}
\caption{Ablation study of each component in SimDistill on the nuScenes val Set. CMD, IMD, MMD (including MMD-F and MMD-P) are the distillation items in SimMMD. GCM denotes the geometry compensation modules used upon the vanilla version of CMD to overcome the modality gap issue. }
\label{tab:ablationSup}
\end{table}
\subsubsection{2.3.2 Ablation Study for CMD}
Based on the vanilla CMD, we exploit GCM with object-aware mask $\mathcal M$ to address the modality gap issue. We conduct a detailed ablation study in Table \ref{tab:CMDablation}. According to the results, solely using GCM is demonstrated very effective to improve the performance of CMD, with an increase of 2.1\% mAP over the vanilla version of CMD. The object-aware mask is also ablated alone, which slightly improves the vanilla CMD with a 0.36\% gain in mAP. However, when the mask is employed together with GCM, the vanilla CMD is largely improved with gains of 3.03\% mAP and 2\% NDS. Thus we use GCM to denote the two combined used techniques to show the advanced ability to overcome modality gap side effects in the main submission. 

\subsubsection{2.3.2 Ablation Study for GCM}
To enable geometry compensation, we resort to using deformable operations to optimize attentive locations and account for geometry misalignment. Both Deformable Convolutions and Deformable Attentions are qualified candidate techniques. Therefore, we conduct an ablation experiment implementing GCM with both deformable operations, as shown in Table \ref{tab:deformchoice}. While the deformable convolution can also attend to various relative regions and improve representations, the evaluation results demonstrate the superiority of multi-head deformable attention operations. In the final version of GCM, we utilize the multi-head deformable attention layers, which optimize attentive locations and corresponding attention weights during training to facilitate geometry correction and improve representations. 

\begin{table}[t]
\footnotesize
\centering
\resizebox{0.36\textwidth}{!}{
\begin{tabular}[t]{c|c|c|c|c|c}
\toprule
& \multicolumn{3}{c|}{CMD}   &\multirow{2}{*}{mAP$\uparrow$}&\multirow{2}{*}{NDS$\uparrow$ }\\
\cline{2-4}
&vanilla&GCM& $\mathcal M$&&\\
\hline
a&$\checkmark$   &&&  36.80  &42.79 \\
b&$\checkmark$   &$\checkmark$&&  38.90  &44.18  \\
c&$\checkmark$   &&$\checkmark$&  37.16  &44.18  \\
d&$\checkmark$   &$\checkmark$&$\checkmark$&  39.83  &44.79  \\
 \bottomrule
\end{tabular}}
\caption{We split CMD into different components including the vanilla version, the version with solely GCM, and the version with solely the object-aware mask $\mathcal M$, to ablate their corresponding effects.}
\label{tab:CMDablation}
\end{table}
\begin{table}[t]
\footnotesize
\centering
\resizebox{0.3\textwidth}{!}{

\begin{tabular}[t]{c|c|c|c}
\toprule
&Methods &mAP$\uparrow$ &NDS$\uparrow$ \\
\midrule
a&DeformAttn &38.90&44.18 \\
b&DeformConv&37.69&43.25 \\
 \bottomrule
\end{tabular}}
\caption{Quantitative comparisons on the nuScenes val Set.}
\label{tab:deformchoice}
\end{table}
\section{Additional Qualitative Analysis}
\begin{figure*}[t]
\begin{center}
  \includegraphics[width=\linewidth]{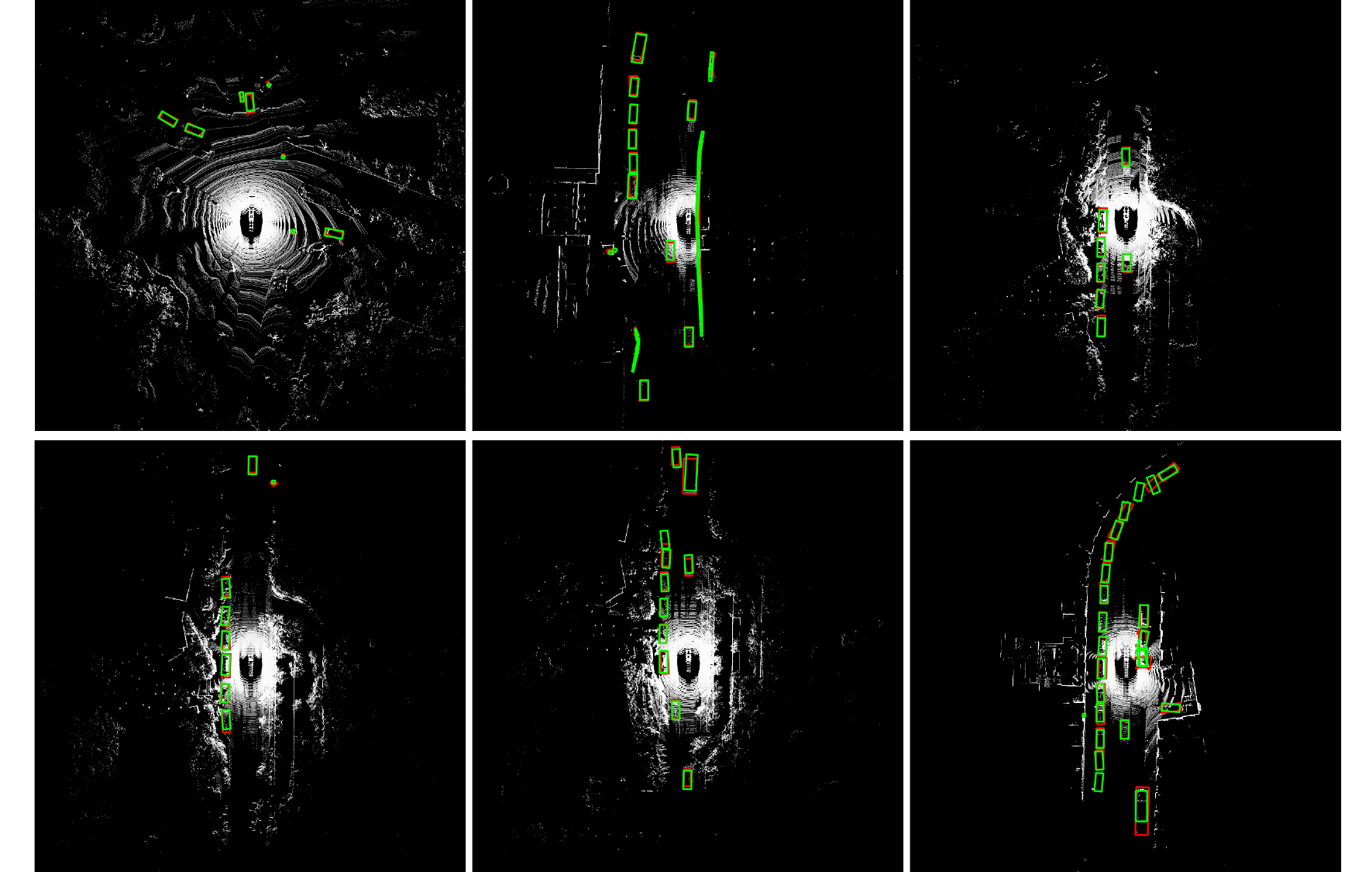}
\end{center}
  \caption{Visualization of the detection results inferred by SimDistill, on LiDAR top view. The green and red boxes represent the prediction and ground truth, respectively.}
\label{fig:lidar}
\end{figure*}
\subsection{Prediction Results}
\begin{figure*}[t]
\begin{center}
  \includegraphics[width=\linewidth]{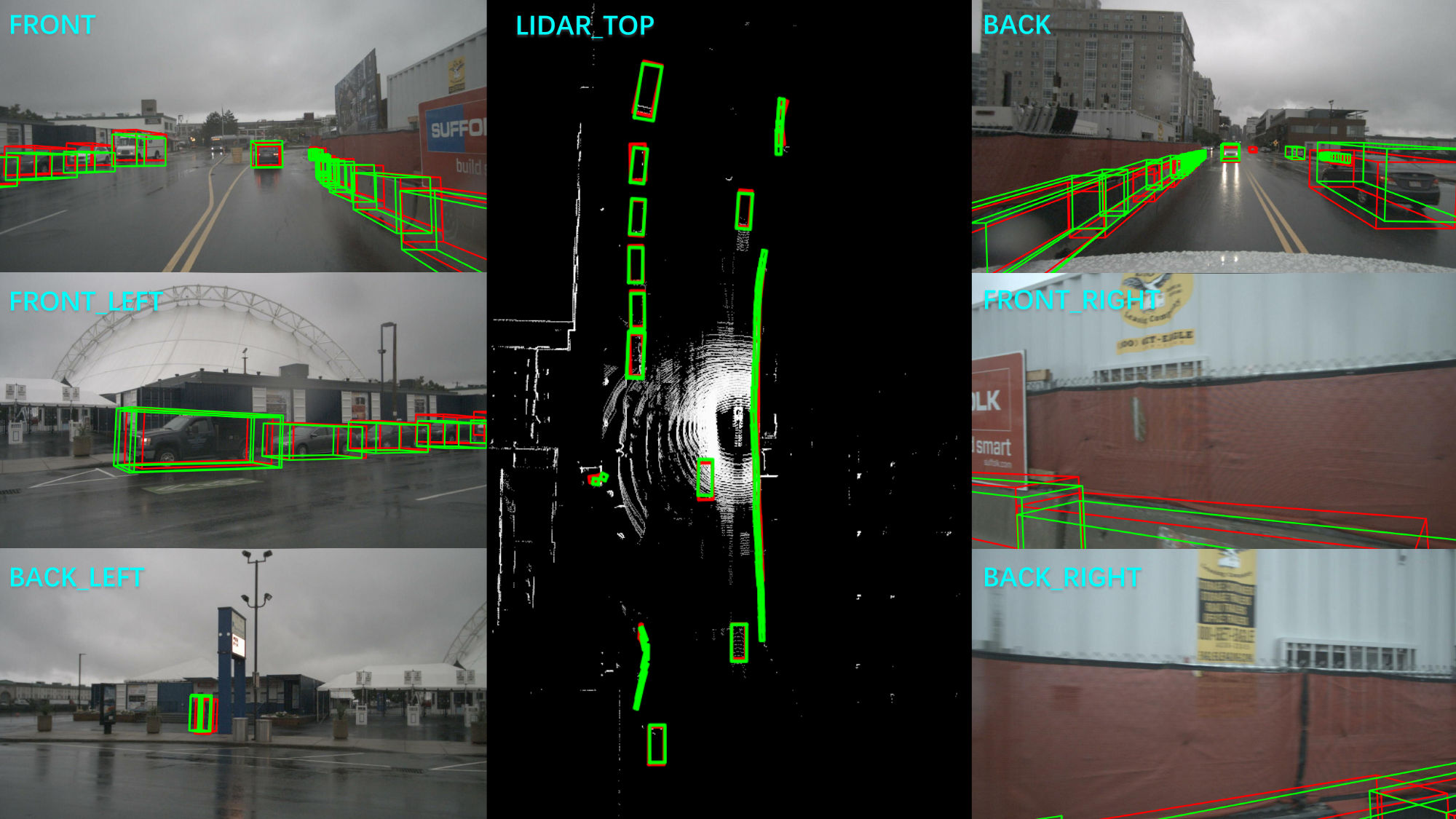}
  \\
  ~\\
  \includegraphics[width=\linewidth]{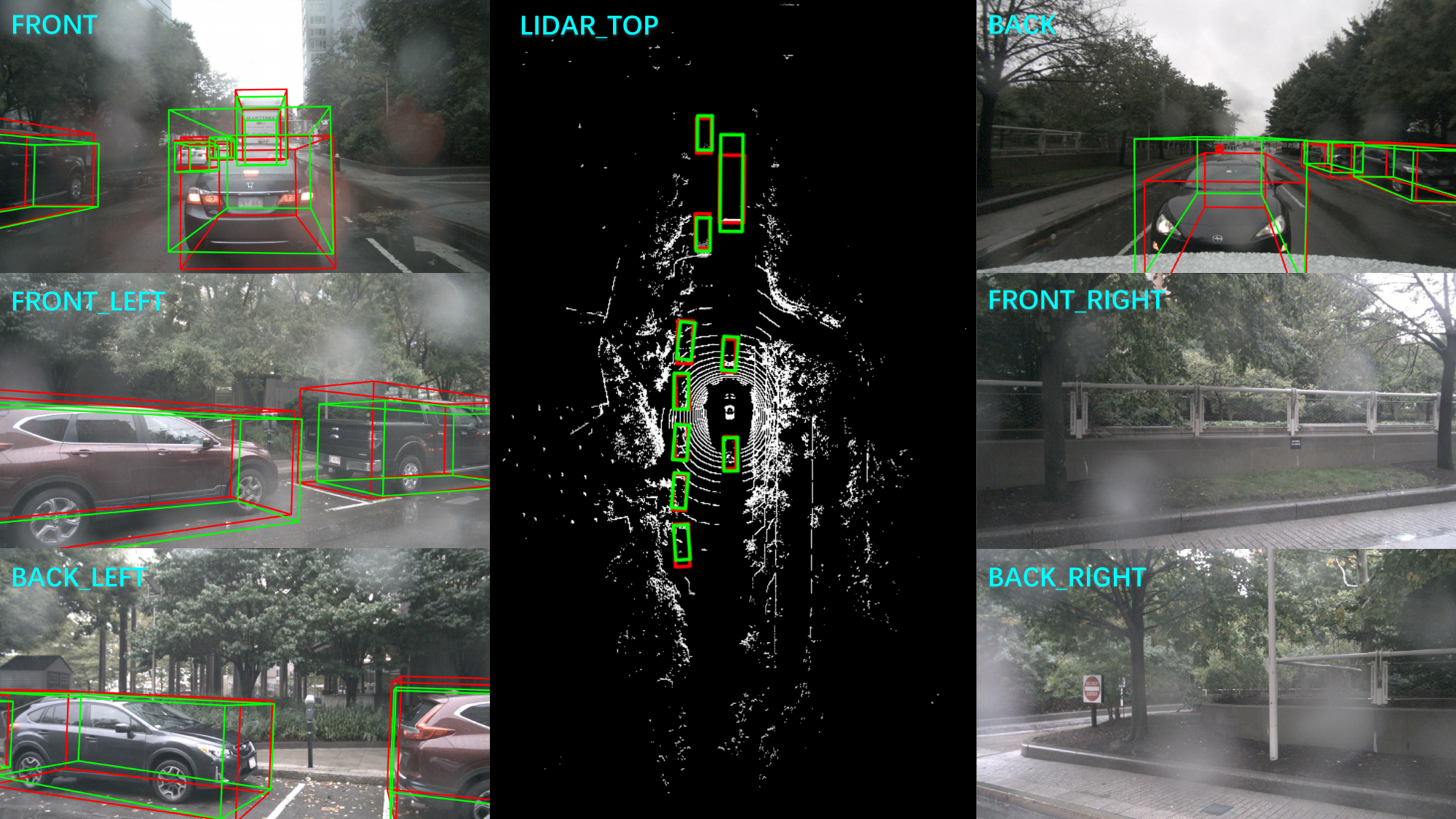}
\end{center}
  \caption{Visualization of the detection results inferred by SimDistill, on both LiDAR top view and surrounded image view. The green and red boxes represent the prediction and ground truth, respectively.}
\label{fig:surround}
\end{figure*}

To demonstrate the performance of SimDistill, we visualize the prediction results in both LiDAR TOP view (Figure \ref{fig:lidar}) and surrounded image view (Figure \ref{fig:surround}). The green and red boxes represent the predictions and the ground truth, respectively. As shown in the figures, our predictions are very close to the ground truth.

\subsection{Feature Map}
To validate the effectiveness of SimDistill, we visualize the BEV feature maps learned in different stages, including the BEV features of the camera branch, the BEV features of the simulated LiDAR branch, and the simulated fusion features.

\begin{figure*}[t]
\begin{center}
  \includegraphics[width=0.9\linewidth]{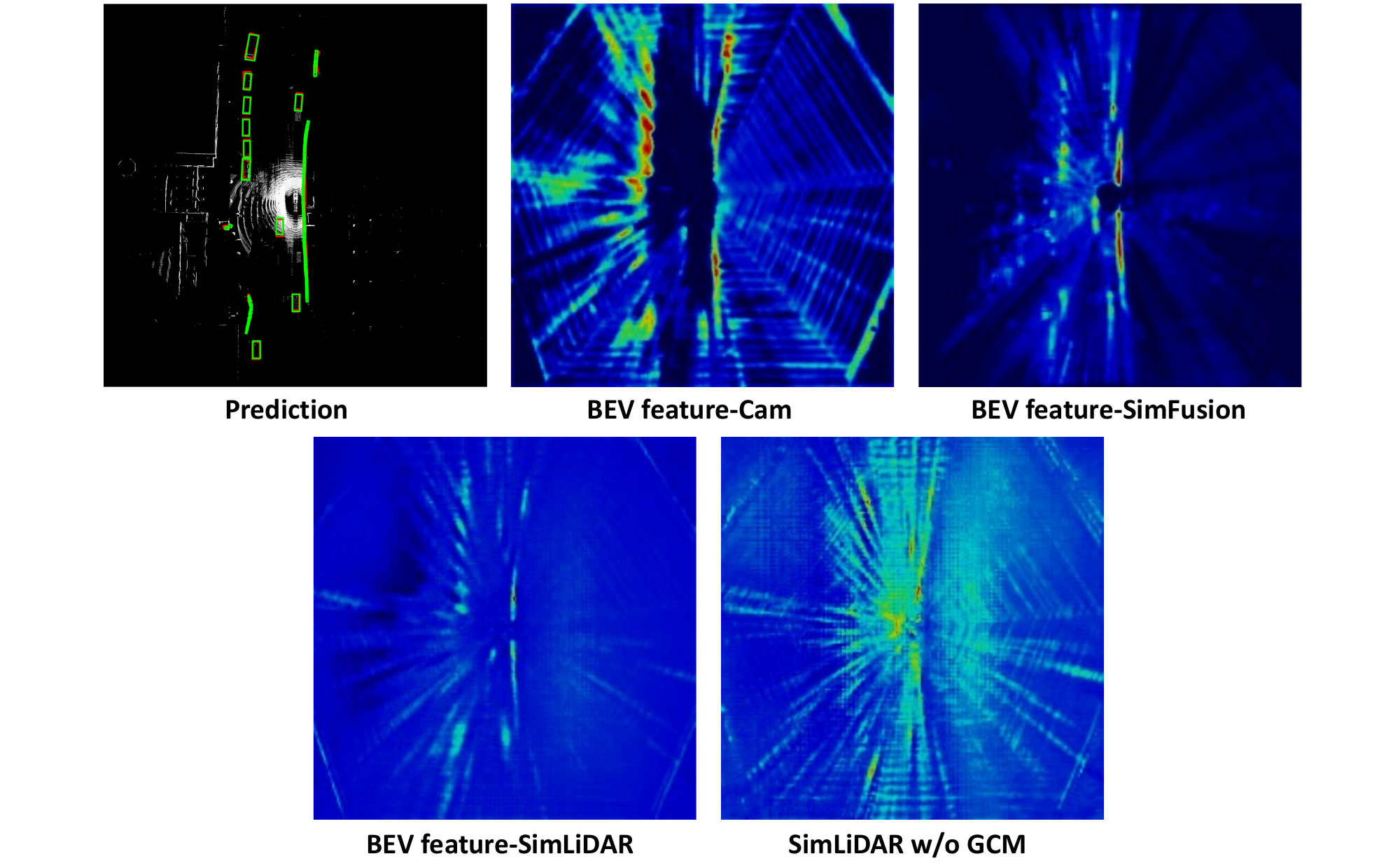}
  \includegraphics[width=0.9\linewidth]{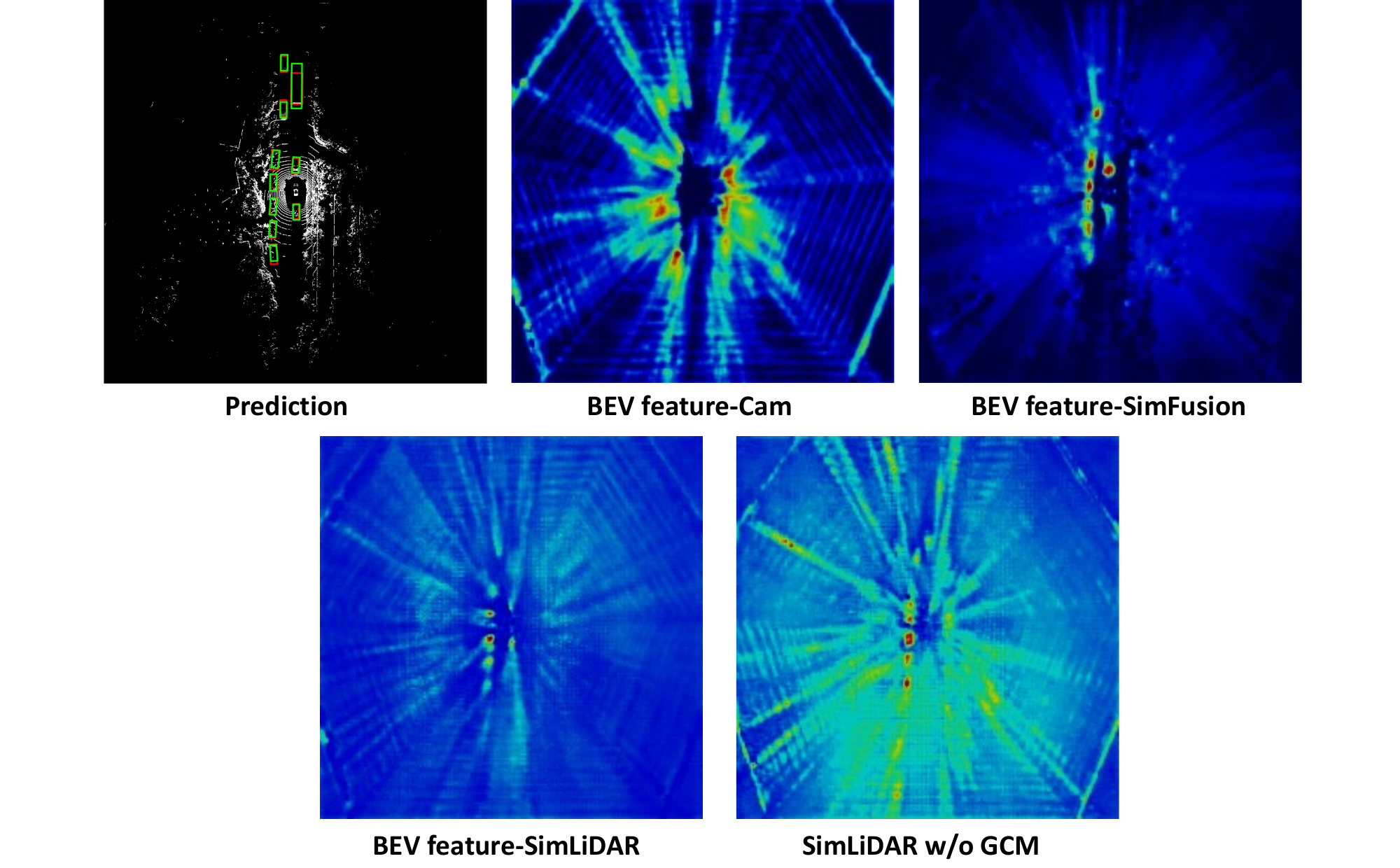}
\end{center}
  \caption{Visualization of BEV feature maps. BEV feature-cam, BEV feature-SimLiDAR and BEV feature-SimFusion represent the BEV features of the camera branch, the simulated LiDAR branch and the fusion ones. The camera branch identifies significant areas and assigns the representation focusing along the camera ray. The simulated LiDAR branch emphasizes more concise regions. The simulated fusion features combine features from both branches to enable more precise identification of object positions that correspond to the prediction and ground truth. SimLiDAR w/o GCM means the BEV features of the simulated LiDAR branch of the model without GCM.}

\label{fig:featuremap}
\end{figure*}

As shown in Figure \ref{fig:featuremap}, the camera branch learns frustum-shape features, identifying the significant area and assigning the representation focus along the camera ray, while the features learned by the simulated LiDAR branch emphasize more concise regions. The simulated fusion features aggregate the features of the two branches and enable more precise identification of object positions that correspond to the prediction and ground truth. The visualization of features demonstrates that the two branches do learn different and complementary information and the fusion features learn more precise and informative representations for detection.

In Figure \ref{fig:featuremap}, we also visualize the simulated LiDAR feature without GCM. Simulated LiDAR features without GCM are similar to the features learned by the camera branch, in the frustum shape, with many redundant context areas for detection. Compared to them, features with GCM can focus more on precise object regions. 

 \begin{figure*}[t]
\begin{center}
  \includegraphics[width=\linewidth]{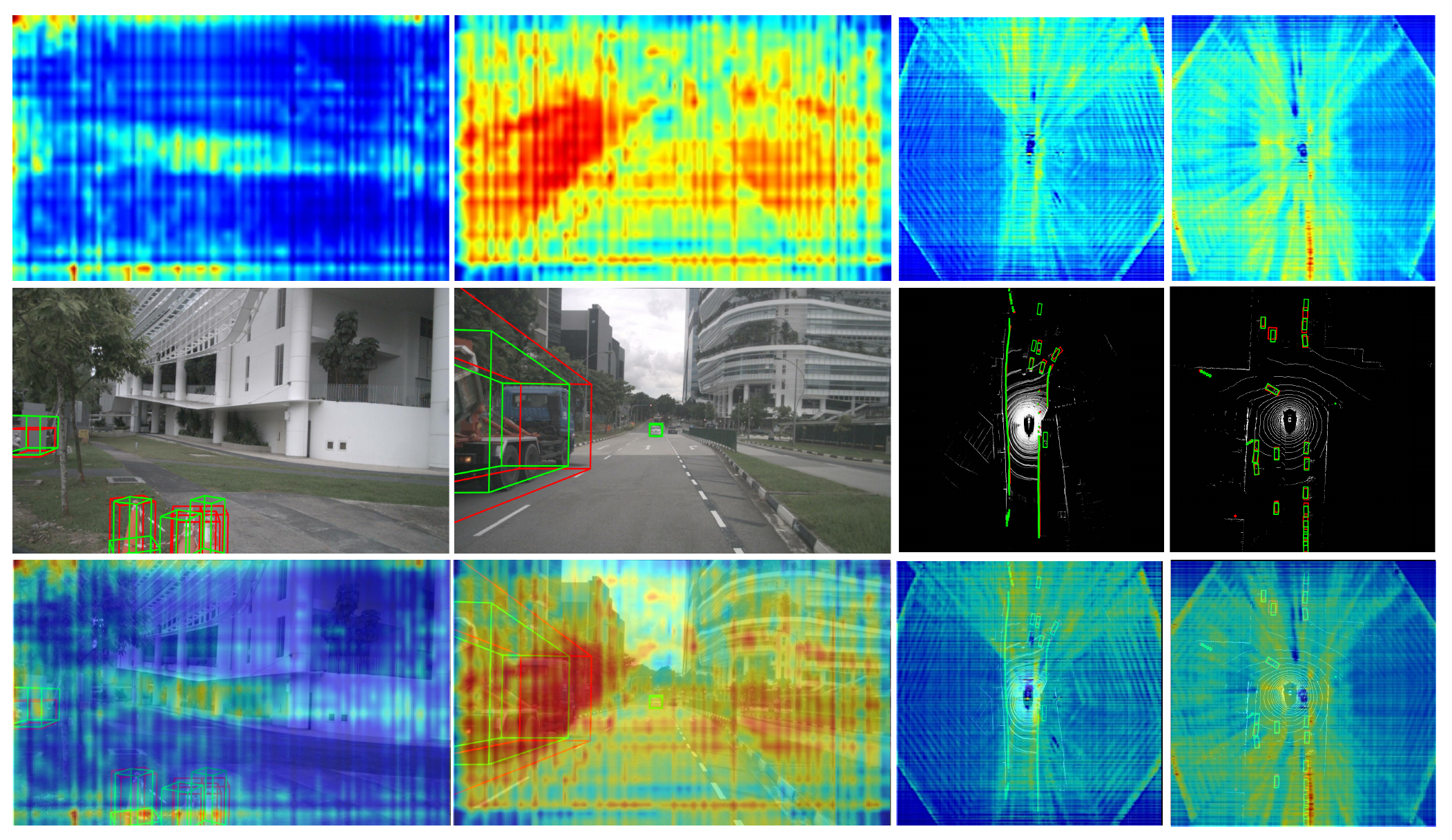}
\end{center}
  \caption{Visualization of the attention maps generated from GCM. The first two columns show the camera-view attention map learned by UV GCM, which mainly attends to the object regions in camera views. The last two columns show the BEV attention map learned by BEV GCM, which primarily focuses on areas with objects in the BEV space.}

\label{fig:gcm}
\end{figure*}
\subsection{Analysis of GCM}
 
To further validate the effect of GCM, we visualize the attention maps in Figure \ref{fig:gcm}. The first two columns show the attention maps learned in UV GCM, where areas with large attention weights align with the object regions in camera views. The last two columns are attention maps learned in the BEV GCM, which also focus primarily on areas with objects in the BEV space.


\bibliography{aaai24}

\end{document}